\documentclass[runningheads]{llncs}
\usepackage{graphicx}
\usepackage{amsmath,amssymb} 
\usepackage{color}
\usepackage[width=122mm,left=12mm,paperwidth=146mm,height=193mm,top=12mm,paperheight=217mm]{geometry}
\usepackage[pagebackref=true,breaklinks=true,letterpaper=true,colorlinks,bookmarks=false]{hyperref}

\usepackage{tablefootnote}
\usepackage{paralist}
\usepackage{epsfig}
\usepackage{graphicx}
\usepackage{caption}
\usepackage{subcaption}
\usepackage{amsmath}
\usepackage{amssymb}
\usepackage{tabularx}
\usepackage{microtype}

\DeclareMathOperator*{\argmax}{argmax}
\DeclareMathOperator*{\argmin}{argmin}

\DeclareMathOperator{\vM}{\mathcal{VM}}

\begin{document}
\renewcommand{\labelitemi}{$\bullet$}
\pagestyle{headings}
\mainmatter
\def\ECCV18SubNumber{770}  

\title{Deep Directional Statistics:~\\Pose Estimation with~\\Uncertainty Quantification} 

\titlerunning{Deep Directional Statistics:~\\Pose Estimation with~\\Uncertainty Quantification}

\authorrunning{S. Prokudin et al.}


\author{Sergey Prokudin\inst{1}
\and Peter Gehler \thanks{This work has been done prior to Peter Gehler joining Amazon.} \inst{2} 
\and Sebastian Nowozin\inst{3}}

\institute{Max Planck Institute for Intelligent Systems, T\"ubingen, Germany,\\
\email{sergey.prokudin@tuebingen.mpg.de}
\and Amazon,  T\"ubingen, Germany
\and Microsoft Research, Cambridge, UK}


\maketitle

\begin{abstract}
Modern deep learning systems  successfully solve many perception tasks such as object pose estimation when the input image is of high quality.
However, in challenging imaging conditions such as on low resolution images or when the image is corrupted by imaging artifacts, current systems degrade considerably in accuracy.
While a loss in performance is unavoidable, we would like our models to quantify their uncertainty in order to achieve robustness against images of varying quality.
Probabilistic deep learning models combine the expressive power of deep learning with uncertainty quantification.
In this paper we propose a novel probabilistic deep learning model for the task of angular regression.
Our model uses \emph{von Mises} distributions to predict a distribution over object pose angle.
Whereas a single von Mises distribution is making strong assumptions about the shape of the distribution, we extend the basic model to predict a mixture of von Mises distributions.
We show how to learn a mixture model using a finite and  \emph{infinite} number of mixture components.
Our model allows for likelihood-based training and efficient inference at test time.
We demonstrate on a number of challenging pose estimation datasets that our model produces calibrated probability predictions and  competitive or superior point estimates  compared to the current state-of-the-art.


\end{abstract}

\section{Introduction}
Estimating object pose is an important building block in systems aiming to understand complex scenes and has a long history in computer vision~\cite{marchand2016pose,murphy2009headpose}.
Whereas early systems achieved low accuracy, recent advances in deep learning and the collection of extensive data sets have led to high performing systems that can be deployed in useful applications~\cite{poirson2016fast,massa2016crafting,beyer2015biternion}.

\begin{figure}[t!]
\centering
    \includegraphics[width=\textwidth]{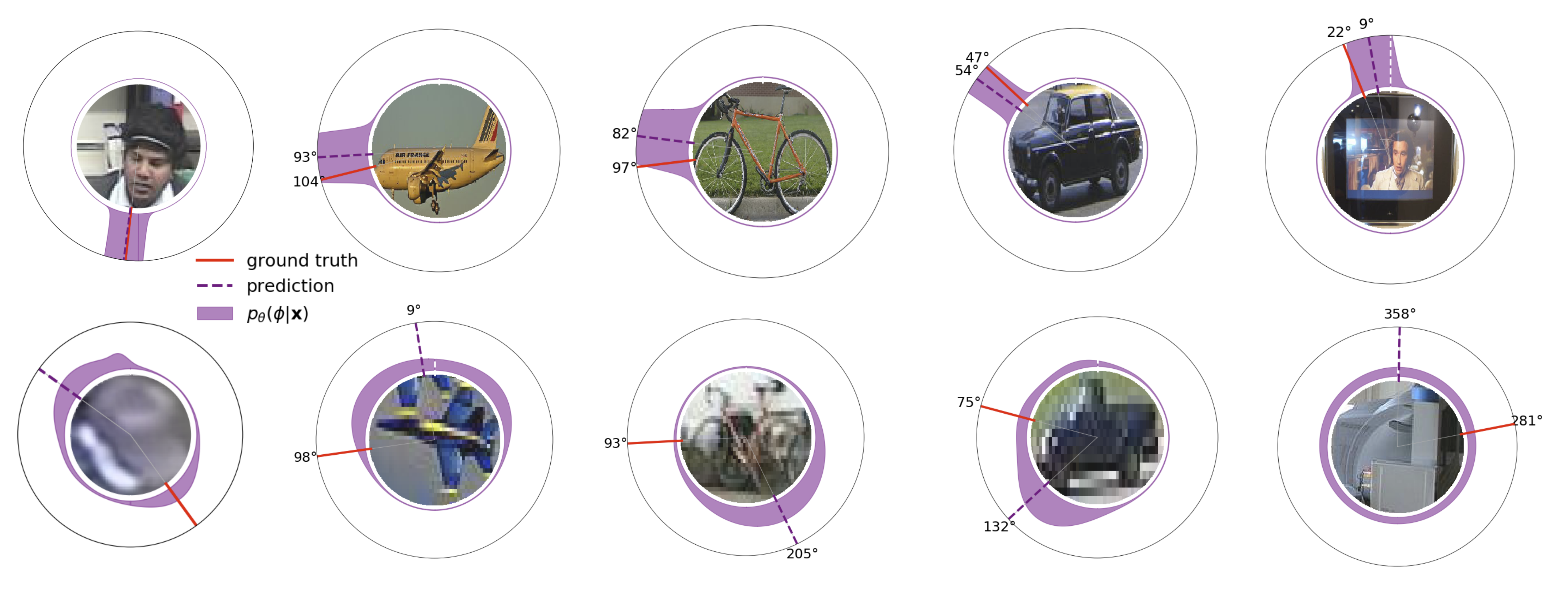}
\caption{%
Our model predicts complex multimodal distributions on the circle (truncated by the outer circle for better viewing). For difficult and ambiguous images our model report high uncertainty (bottom row). Pose estimation predictions (pan angle) on images from IDIAP, TownCentre and PASCAL3D+ datasets. }
\label{fig:fig1}
\end{figure}

However, the reliability of object pose regression depends on the quality of the image provided to the system. Key challenges are low-resolution due to distance of an object to the camera, blur due to motion of the camera or the object, and sensor noise in case of poorly lit scenes (see Figure \ref{fig:fig1}).

We would like to predict object pose in a way that captures uncertainty.
\emph{Probability} is the right way to capture the uncertainty~\cite{berger1980bayesian} and in this paper we therefore propose a novel model for object pose regression whose predictions are fully probabilistic. Figure \ref{fig:fig1} depicts an output of the proposed system.
Moreover, instead of assuming a fixed form for the predictive density we allow for very flexible distributions, specified by a deep neural network.

The value of quantified uncertainty in the form of probabilistic predictions is two-fold:
\emph{first}, a high prediction uncertainty is a robust way to diagnose poor inputs to the system;
\emph{second}, given accurate probabilities we can summarize them to improved point estimates using Bayesian decision theory.

More generally, accurate representation of uncertainty is especially important in case a computer vision system becomes part of a larger system, such as when providing an input signal for an autonomous control system.
If uncertainty is not well-calibrated, or---even worse---is not taken into account at all, then the consequences of decisions made by the system cannot be accurately assessed, resulting in poor decisions at best, and dangerous actions at worst.

In the following we present our method and make the following contributions:
\begin{itemize}
\item We demonstrate the importance of probabilistic regression on the application of object pose estimation;
\item We propose a novel efficient probabilistic deep learning model for the task of circular regression;
\item We show on a number of challenging pose estimation datasets (including PASCAL 3D+ benchmark \cite{xiang2014beyond}) that the proposed probabilistic method outperforms purely discriminative approaches in terms of predictive likelihood and show competitive performance in terms of angular deviation losses classically used for the tasks.
\end{itemize}

\section{Related Work}
Estimation of object orientation arises in different applications and in this paper we focus on the two most prominent tasks: head pose estimation and object class orientation estimation. Although those tasks are closely related, they have been studied mostly in separation, with methods applied to exclusively one of them. We will therefore discuss them separately, despite the fact that our model applies to both tasks. We present results for our methods on the standard benchmarks from both domains. 

\emph{Head pose estimation} has been a subject of extensive research in computer vision for a long time ~\cite{siriteerakul2012advance,murphy2009headpose} and the existing systems vary greatly in terms of feature representation and proposed classifiers.
The input to pose estimation systems typically consists of 2D head images~\cite{idiaphead,gourier2004estimating,demirkus2014robust}, and often one has to cope with low resolution images~\cite{murphy2007head,fisher2005caviar,benfold2011unsupervised,siriteerakul2012advance}. 
Additional modalities such as depth~\cite{fanelli2011real} and motion \cite{benfold2011unsupervised,chamveha2013head} information has been exploited and provides useful cues. However, these are not always available.
Also, information about the full body image could be used for joint head and body pose prediction~\cite{chen2012we,Flohr2015APF,osadchy2007synergistic}. Notably the work of~\cite{Flohr2015APF} also promotes a probabilistic view and fuse body and head orientation within a tracking framework. 
Finally, the output of facial landmarks can be used as an intermediate step~\cite{dantone2012real,zhu2012face}. 

Existing head pose estimation models are diverse and include manifold learning approaches~\cite{lu2013headpose,huang2011supervised,tosato2013characterizing,benabdelkader2010robust}, energy-based models \cite{osadchy2007synergistic}, linear regression based on HOG features \cite{geng2014headpose}, regression trees \cite{fanelli2011real,kazemi2014one} and convolutional neural networks \cite{beyer2015biternion}. 
A number of probabilistic methods for head pose analysis exist in the literature~\cite{ba2004probabilistic,demirkus2014probabilistic,Flohr2015APF}, but none of them combine probabilistic framework with learnable hierarchical feature representations from deep CNN architectures.
At the same time, deep probabilistic models have shown an advantage over purely discriminative models in other computer vision tasks, e.g., depth estimation \cite{kendall2017uncertainties}. 
To the best of our knowledge, our work is the first to utilize deep probabilistic approach to angular orientation regression task. 

An early dataset for estimating the \emph{object rotation for general object classes} was proposed in~\cite{savarese07iccv} along with an early benchmark set. Over the years the complexity of data increased, from object rotation ~\cite{savarese07iccv} and images of cars in different orientations~\cite{ozuysal2009epflcars} to Pascal3D~\cite{xiang_wacv14}. The work of~\cite{xiang_wacv14} then assigned a separate Deformable Part Model (DPM) component to a discrete set of viewpoints. The work of~\cite{pepik12eccv,pepik12teaching} then proposed different 3D DPM extensions which allowed viewpoint esimation as integral part of the model. However, both~\cite{pepik12eccv} and ~\cite{pepik12teaching} and do not predict a continuous angular estimate but only a discrete number of bins. 

More recent versions make use of CNN models but still do not take a probabilistic approach~\cite{poirson2016fast,massa2016crafting}. The work of~\cite{su2015render} investigates the use of a synthetic rendering pipeline to overcome the scarcity of detailed training data. The addition of synthetic and real examples allows them to outperform previous results. The model in \cite{su2015render} predicts angles, and constructs a loss function that penalizes geodesic and $\ell_1$ distance. In this work we advocate the use of likelihood estimation as a principled probabilistic training objective. The recent work of ~\cite{tulsiani2015viewpoints} draws a connection between viewpoints and object keypoints. The viewpoint estimation is however again, framed as a classification problem in terms of Euler angles to obtain a rotation matrix from a canonical viewpoint. 

Many works phrase angular prediction as a classification problem~\cite{poirson2016fast,tulsiani2015viewpoints,su2015render} which always limits the granularity of the prediction and also requires the design of a loss function and a means to select the number of discrete labels. A benefit of a classification model is that components like softmax loss can be re-used and also interpreted as an uncertainty estimate. In contrast, our model mitigate this problem: the likelihood principle suggests a direct way to train parameters, moreover ours is the only model in this class that conveys an uncertainty estimate.

\section{Review of Biternion Networks}
\label{sec:biternion_net}

We build on the Biternion networks method for pose estimation from~\cite{beyer2015biternion} and briefly review the basic ideas here.
Biternion networks regress angular data and currently define the state-of-the-art model for a number of challenging  head pose estimation datasets. 

A key problem is to regress angular orientations which is periodic and prevents a straight-forward application of standard regression methods, including CNN models with common loss functions.
Consider a ground truth value of $0^{\circ}$, then both predictions $1^{\circ}$ and $359^{\circ}$ should result in the same absolute loss.
Applying the $\mod$ operator is no simple fix to this problem, since it results in a discontinuous loss function that complicates the optimization.
A loss function needs to be defined to cope with this discontinuity of the target value.
Biternion networks overcome this difficulty by using a different parameterization of angles and the cosine loss function between angles.

\subsection{Biternion Representation} 
\label{subsec:biternion_net}   

Beyer et al.~\cite{beyer2015biternion} propose an alternative representation of an angle $\phi$ using the two-dimensional sine and cosine components $\vec{y} = (\cos{\phi}, \sin{\phi})$. 

%
%
This {\em biternion representation} is inspired by quaternions, which are popular in computer graphics systems.
It is easy to predict a $(\cos, \sin)$ pair with a fully-connected layer followed by a normalization layer, that is, 
\begin{align}
    f_{BT}(\vec{x}; \vec{W}, \vec{b})= \frac{\vec{W}\vec{x}+\vec{b}}{|| \vec{W}\vec{x}+\vec{b} ||}.
    \label{eq:biternion}
\end{align}
A Biternion network is then a convolutional neural network with a layer (\ref{eq:biternion}) as the final operation, outputting a two-dimensional vector $\vec{y_{pred}}$. We use VGG-style network \cite{simonyan2014very} and InceptionResNet \cite{szegedy2017inception} networks in our experiments and  provide a detailed description of the network architecture in Section \ref{subsec:net_arch}. 
Given recent developments in network architectures it is likely that different network topologies may perform better than selected backbones. We leave this for future work, our contributions are orthogonal to the choice of the basis model.


\subsection{Cosine loss function} 
\label{subsec:cosine_loss} 
The cosine distance is chosen in~\cite{beyer2015biternion} as a natural candidate to measure the difference between the predicted and ground truth Biternion vectors. It reads
\begin{eqnarray}
L_{cos}(\vec{y_{pred}} , \vec{y_{true}})
& = & 1 - \frac{\vec{y_{pred}} \cdot \vec{y_{true}}}{ ||\vec{y_{pred}}|| \cdot ||\vec{y_{true}}||}  =  1 - \vec{y_{pred}} \cdot \vec{y_{true}},
\label{eq:cosine_loss}
\end{eqnarray}
where the last equality is due to $|| \vec{y} || = \cos^2{\phi} + \sin^2{\phi} = 1$.

The combination of a Biternion angle representation and a cosine loss solves the problems of regressing angular values, allowing for a flexible deep network with angular output.
We take this state-of-the-art model and generalize it into a family of probabilistic models of gradually more flexibility.

\section{Probabilistic models of circular data.}
\label{sec:von_mises}

\begin{figure}[t!]
\centering
    \includegraphics[width=\textwidth]{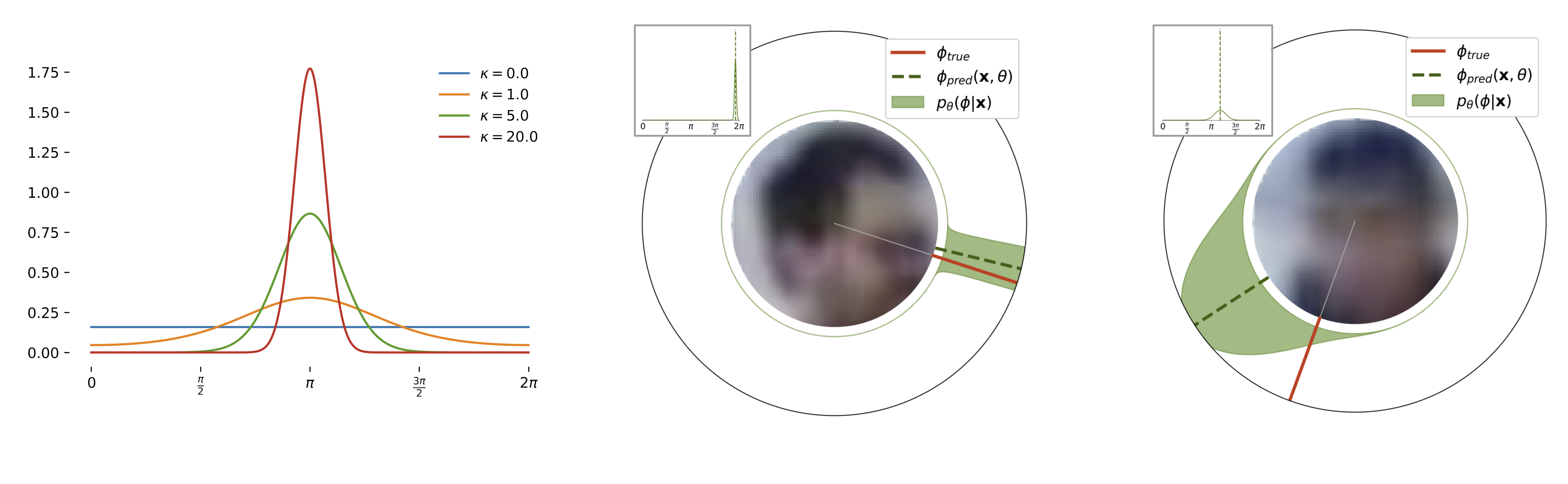}
    \caption{Left: examples of the von Mises probability density function for different concentration parameters $\kappa$. Center, right: predicted $\vM$ distributions for two images from the CAVIAR dataset. We plot the predicted density on the viewing circle. For comparison we also include the 2D plot (better visible in zoomed pdf version). The distribution on the center image is very certain, the one on the right more uncertain about the viewing angle.}
    \label{fig:adjusted_densities}
\end{figure}

The von Mises (vM) distribution is the basic building block of our probabilistic framework for circular data~\cite{mardia2009directional}. We continue with a brief formal definition and in Section~\ref{subsec:vm_from_biternion} describe a simple way to convert the output of Biternion networks into a $\vM$ density, that does not require any network architecture change or re-training as it requires only selection of the model variance. 
We will then use this approach as a baseline for more advanced probabilistic models. 
Section~\ref{subsec:vm_loglikelihood} slightly extends the original Biternion network by introducing an additional network output unit that models uncertainty of our angle estimation and allows optimization for the log-likelihood of the $\vM$ distribution. 

The von Mises distribution $\vM (\mu, \kappa)$ is a close approximation of a normal distribution on the unit circle. Its probability density function is
\begin{align}
     p(\phi ; \mu, \kappa) = \frac{\exp{(\kappa \cos{(\phi - \mu)})}}{2\pi I_0(\kappa)},
    \label{eq:von_mises}
\end{align}
where $\mu\in[0,2\pi)$ is the mean value,  $\kappa\in\mathbb{R}_+$ is a measure of concentration (a reciprocal measure of dispersion, so 1/$\kappa$ is analogous to $\sigma^2$ in a normal distribution), and $I_0(\kappa)$ is the modified Bessel function of order 0. We show examples of $\vM$-distributions with $\mu=\pi$ and varying $\kappa$ values in Figure~\ref{fig:adjusted_densities} (left).

%
%

\subsection{Von Mises Biternion Networks}
\label{subsec:vm_from_biternion} 

A conceptually simple way to turn the Biternion networks from Section~\ref{sec:biternion_net} into a probabilistic model is to take its predicted value as the center value of the $\vM$ distribution,
\begin{align}
     p_{\theta}(\phi | \vec{x}; \kappa) = \frac{\exp{(\kappa \cos{(\phi - \mu_{\theta}(\vec{x}))})}}{2\pi I_0(\kappa)},
    \label{eq:von_mises_fixed_biternion}
\end{align}
where $\vec{x}$ is an input image, $\theta$ are parameters of the network, and $\mu_{\theta}(\vec{x})$ is the network output. In order to arrive at a probability distribution we may regard $\kappa > 0$ as a hyper-parameter.
For fixed network parameters $\theta$ we can select $\kappa$ by maximizing the log-likelihood of the observed data,
\begin{align}
    \kappa^* =
    \argmax_{\kappa} \sum_{i=1}^N{\log p_{\theta}(\phi^{(i)}| \vec{x}^{(i)}; \kappa) }.
\end{align}
The model~(\ref{eq:von_mises_fixed_biternion}) with $\kappa^*$ will serve as the simplest probabilistic baseline in our comparisons.

\subsection{Maximizing the von Mises Log-likelihood}

\label{subsec:vm_loglikelihood}
Using a single scalar $\kappa$ for every possible input in the model~(\ref{eq:von_mises_fixed_biternion}) is clearly a restrictive assumption: model certainty should depend on factors such as image quality, light conditions, etc. 
For example, Figure~\ref{fig:adjusted_densities} (center, right) depicts  two low resolution images from a surveillance camera that are part of the CAVIAR dataset~\cite{fisher2005caviar}. 
In the left image facial features like eyes and ears are distinguishable which allows a model to be more certain when compared to the more blurry image on the right.

We therefore extend the simple model by replacing the single constant $\kappa$ with a function $\kappa_{\theta}(\vec{x})$, predicted by the Biternion network,
\begin{align}
     p_{\theta}(\phi | \vec{x}) = \frac{\exp{(\kappa_{\theta}(\vec{x}) \cos{(\phi - \mu_{\theta}(\vec{x}))})}}{2\pi I_0(\kappa_{\theta}(\vec{x}))}.
    \label{eq:von_mises_full_biternion}
\end{align}

We train~(\ref{eq:von_mises_full_biternion}) by maximizing the log-likelihood of the data,
\begin{equation}
\log{\mathcal{L}(\theta | \vec{X}, \Phi)}
=  \sum_{i=1}^N{\kappa_{\theta}(\vec{x}^{(i)})\cos{(\phi^{(i)} - \mu_{\theta}(\vec{x}^{(i)}))}}
- \sum_{i=1}^N{\log 2 \pi I_0(\kappa_{\theta}(\vec{x}^{(i)}))}.
    \label{eq:von_mises_likelihood}
\end{equation}
Note that when $\kappa$ is held constant in~\eqref{eq:von_mises_likelihood}, the second sum in
$\log{\mathcal{L}(\theta | \vec{X}, \Phi)}$ is constant and therefore we recover the
Biternion cosine objective~\eqref{eq:cosine_loss} up to constants $C_1$, $C_2$,
\begin{equation*}
\log{\mathcal{L}(\theta| \vec{\vec{X}}, \Phi, \kappa)}
= C_1 \sum_{i=1}^N{\cos{\big(\phi^{(i)} - \mu_{\theta}(\vec{x}^{(i)})\big)}} + C_2.
\end{equation*}
The sum has the equivalent form,
\begin{eqnarray}
\sum_{i=1}^N{\cos{\big(\phi^{(i)} - \mu_{\theta}(\vec{x}^{(i)})\big)}}
& = & \sum_{i=1}^N{\big[\cos{\phi^{(i)}} \cos{\mu_{\theta}(\vec{x}^{(i)})} + \sin{\phi^{(i)}}\sin{\mu_{\theta}(\vec{x}^{(i)})} \big]}
\\
& = & \sum_{i=1}^N{\vec{y}_{\phi^{(i)}} \cdot \vec{y}_{\mu_{\theta}(\vec{x}^{(i)})} },
\label{eq:cosine_likelihood_equality}
\end{eqnarray}

\begin{figure*}[t!]
\centering
\includegraphics[width=\textwidth]{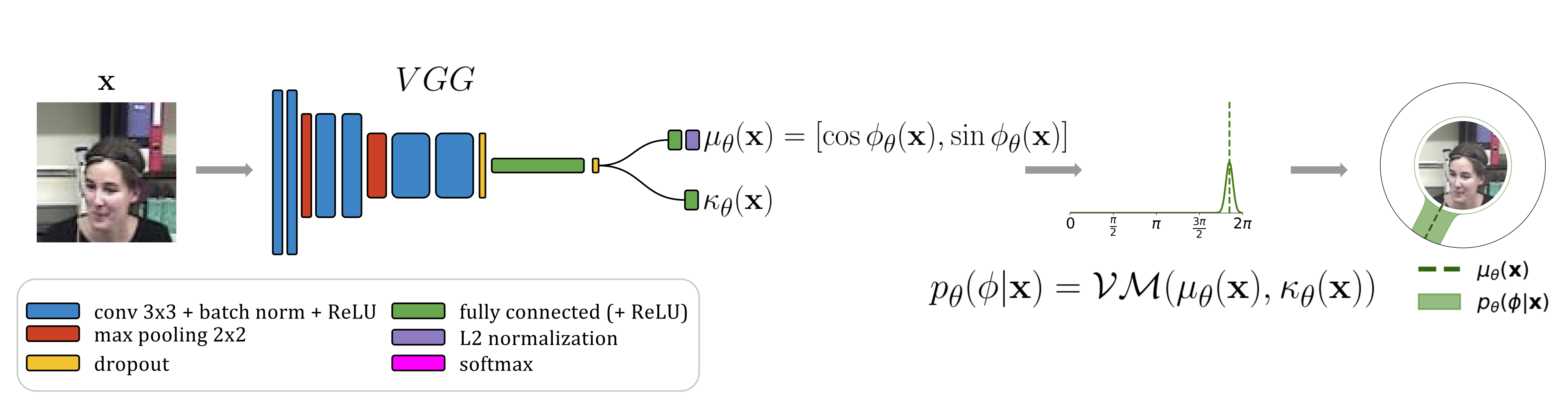}
\caption{\textbf{The single mode von Mises model (VGG backbone variation)}. A BiternionVGG network regresses both mean and concentration parameter of a single vM distribution.}
\label{fig:single_vm}
\end{figure*}

%
where $\vec{y}_{\phi} {=} (\cos{\phi}, \sin{\phi})$ is a Biternion representation of an angle.
Note, that the above derivation shows that the loss function in~\cite{beyer2015biternion} corresponds to optimizing
the von Mises log-likelihood for the fixed value of $\kappa=1$. This offers an interpretation of Biternion networks as a probabilistic model.

The additional degree of freedom to learn $\kappa_{\theta}(\vec{x})$ as a function of $\vec{x}$ allows us to capture the desired image-dependent uncertainty as can be seen in Figure~\ref{fig:adjusted_densities}.

However, like the Gaussian distribution the von Mises distribution makes a specific assumption regarding the shape of the density.  We now show how to overcome this limitation by using a mixture of von Mises distributions.


\section{Mixture of von Mises Distributions} 
\label{sec:vm_mixture}

The model described in Section~\ref{subsec:vm_loglikelihood} is only unimodal and can not capture ambiguities in the image. However, in case of blurry images like the ones in Figure~\ref{fig:adjusted_densities} we could be interested in distributing the mass around a few potential high probability hypotheses, for example, the model could predict that a person is looking sideways, but could not determine the direction, left or right, with certainty.
In this section we present two models that are able to capture multimodal beliefs while retaining a calibrated uncertainty measure.

\subsection{Finite Mixture of von Mises Distributions}
\label{subsec:finite_vm_mixture}

One common way to generate complex distributions is to sum multiple distributions into a {\em mixture distribution}.
We introduce $K$ different component distributions and a $K$-dimensional probability vector representing the mixture weights.
Each component is a simple von Mises distribution. We can then represent our density function as
\begin{equation}
p_{\theta}(\phi | \vec{x})
= \sum_{j=1}^{K}{\pi_j(\vec{x}, \theta) \, p_j(\phi| \vec{x}, \theta )},
\end{equation}
where $p_j(\phi| \vec{x}, \theta) = \vM (\phi | \mu_j, \kappa_j)$ for $j=1,\dots,K$ are the $K$ component distributions and the mixture weights are $\pi_j(\vec{x}, \theta)$ so that $\sum_j \pi_j(\vec{x},\theta) = 1$.
We denote all parameters with the vector $\theta$, it contains component-specific parameters as well as parameters shared across all components.

\begin{figure*}[t!]
\centering
    \includegraphics[width=\textwidth]{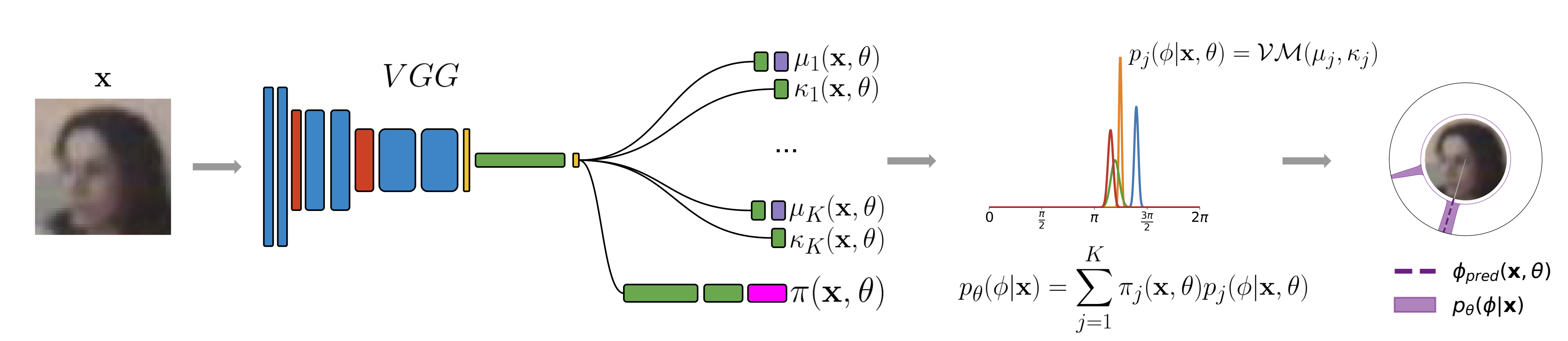}
    \caption{\textbf{The finite $\vM$ mixture model}. A VGG network predicts $K$ mean and concentration values and the mixture coefficients $\pi$. This allows to capture multimodality in the output.}
     \label{fig:finite_vm_mixture}
\end{figure*}

%


To predict the mixture in a neural network framework, we need $K \times 3$ output units for modeling all von Mises component parameters (two for modeling the Biternion representation of the mean, $\mu_j(\vec{x},\theta)$ and one for the $\kappa_j(\vec{x},\theta)$ value), as well as $K$ units for the probability vector $\pi_j(\vec{x},\theta)$, defined by taking the \texttt{softmax} operation to get a positive mixture weights.





The finite von Mises density model then takes form
\begin{equation}
p_{\theta}(\phi | \vec{x}) = \sum_{j=1}^{K}{\pi_j(\vec{x}, \theta) \,
    \frac{\exp{\Big(\kappa_{j}(\vec{x}, \theta) \cos{\big(\phi - \mu_{j}(\vec{x}, \theta)\big)}\Big)}}{2\pi I_0\big(\kappa_{j}(\vec{x}, \theta)\big)}}.
    \label{eq:von_mises_full_finite_mixture}
\end{equation}
Similarly to the single von Mises model, we can train by maximizing the log-likelihood of the observed data, $\sum_{i=1}^N \log p_{\theta}(\phi^{(i)} | \vec{x}^{(i)})$.
We show an overview of the model in Figure~\ref{fig:finite_vm_mixture}.

\subsection{Infinite Mixture (CVAE)}
\label{subsec:infinite_cvae}

\begin{figure*}[t!]
\centering
\includegraphics[width=\textwidth]{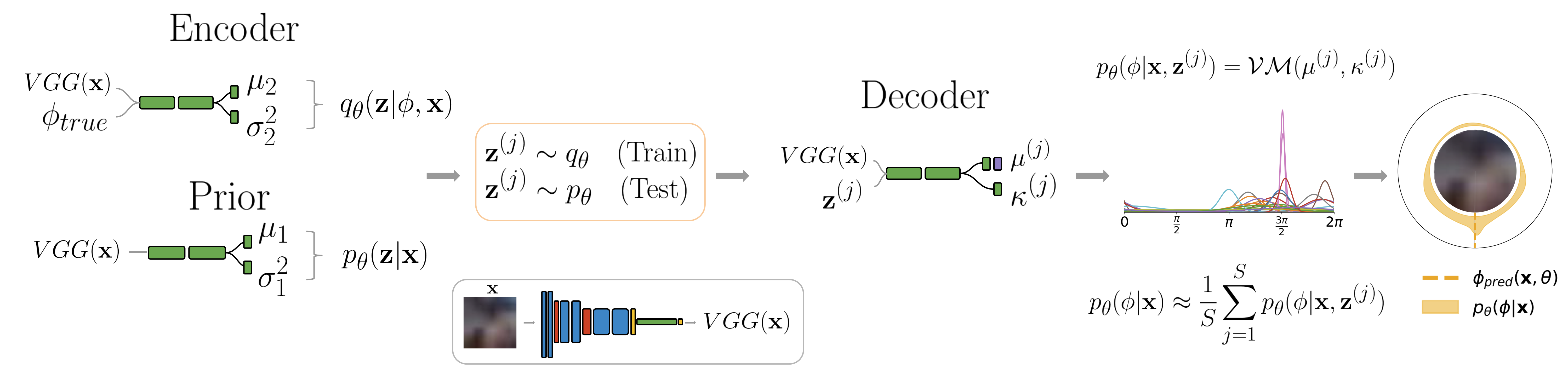}
\caption{\textbf{The infinite mixture model (CVAE)}. An encoder network predicts a distribution $q(\vec{z}|\vec{x})$ over latent variables $\vec{z}$, and a decoder network $p(\phi|\vec{x},\vec{z})$ defines individual mixture components. Integrating over $\vec{z}$ yields an infinite mixture of von Mises distributions.
In practice we approximate this integration using a finite number of Monte Carlo samples $\vec{z}^{(j)} \sim q(\vec{z}|\vec{x})$.}
\label{fig:cvae}
\end{figure*}

To extend the model from a finite to an infinite mixture model, we follow the variational autoencoder (VAE) approach~\cite{kingma2013auto,rezende2014stochastic}, and introduce a vector-valued latent variable $\vec{z}$.
The resulting model is depicted in Figure~\ref{fig:cvae}.
The continuous latent variable becomes the input to a decoder network $p(\phi|\vec{x},\vec{z})$ which predicts the parameters---mean and concentration---of a single von Mises component.
We define our density function as the infinite sum (integral) over all latent variable choices, weighted by a learned distribution $p(\vec{z}|\vec{x})$,
\begin{align}
p_{\theta}(\phi | \vec{x}) = \int{p(\phi|\vec{x},\vec{z}) \, p(\vec{z}|\vec{x}) d\vec{z}},
\label{eq:von_mises_full_infinite_mixture}
\end{align}
where $ p_{\theta}(\phi| \vec{x}, \vec{z}) = \vM (\mu(\vec{x}, \theta), \kappa(\vec{x}, \theta))$, and $p_{\theta}(\vec{z}| \vec{x}) =  \mathcal{N}(\mu_1(\vec{x}, \theta), \sigma_1^2(\vec{x}, \theta))$.
The log-likelihood $\log p_{\theta}(\phi | \vec{x})$ for this model is not longer tractable, preventing simple maximum likelihood training.
Instead we use the variational autoencoder framework of~\cite{kingma2013auto,rezende2014stochastic} in the form of the conditional VAE (CVAE)~\cite{sohn2015learning}.
The CVAE formulation uses an auxiliary \emph{variational} density
$q_{\theta}(\vec{z}|\vec{x}, \phi) =  \mathcal{N}(\mu_2(\vec{x}, \phi, \theta),\\ \sigma_2^2(\vec{x}, \phi, \theta))$ and instead of the log-likelihood optimizes a
{\em variational lower bound},
\begin{eqnarray}
\log{ p_{\theta}(\phi | \vec{\vec{x}})}
& = & 
\log{\int{ p_{\theta}(\phi|\vec{x},\vec{z}) \, p_{\theta}(\vec{z}|\vec{x}) d\vec{z}}}\\
& \geq &
\mathbb{E}_{z \sim q_{\theta}(\vec{z}|\vec{x},\phi)}\left[
    \log \frac{p_{\theta}(\phi | \vec{x},\vec{z}) \, p_{\theta}(\vec{z}|\vec{x})}{
        q_{\theta}(\vec{z}|\vec{x},\phi)}
\right] =: \mathcal{L}_{\textrm{ELBO}}(\theta | \vec{x},\phi).\label{eq:elbo-exp}
\end{eqnarray}
%
We refer  to~\cite{kingma2013auto,sohn2015learning,doersch2016tutorial,rezende2014stochastic} for more details on VAEs.

The CVAE model is composed of multiple deep neural networks:
an {\em encoder network} $q_{\theta}(\vec{z}|\vec{x}, \phi)$,
a {\em conditional prior network} $ p_{\theta}(\vec{z}| \vec{x})$, and
a {\em decoder network} $ p_{\theta}(\phi| \vec{x}, \vec{z})$.
Like before, we use $\theta$ to denote the entirety of trainable parameters of all three model components.
We show an overview of the model in Figure~\ref{fig:cvae}.
The model is trained by maximizing the variational lower bound~\eqref{eq:elbo-exp} over the training set $(\vec{X},\Phi)$,
where
$\vec{X}=(\vec{x}^{(1)},\dots,\vec{x}^{(N)})$ are the images and
$\Phi = (\phi^{(1)},\dots,\phi^{(N)})$ are the ground truth angles.
We maximize
\begin{eqnarray}
\hat{\mathcal{L}}_{\textrm{CVAE}}(\theta | \vec{X}, \Phi)
& = & \frac{1}{N} \sum_{i=1}^N \hat{\mathcal{L}}_{\textrm{ELBO}}(\theta | \vec{x}^{(i)}, \phi^{(i)}),
\label{eq:cvae_objective}
\end{eqnarray}
%
where we use $\hat{\mathcal{L}}_{\textrm{ELBO}}$ to denote the Monte Carlo approximation to~\eqref{eq:elbo-exp} using $S$ samples.
We can optimize~\eqref{eq:cvae_objective} efficiently using stochastic gradient descent.

To evaluate the log-likelihood during testing, we use the importance-weighted sampling technique proposed in~\cite{burda2015iwae} to derive a stronger bound on the marginal likelihood,
\begin{eqnarray}
\log p_{\theta}(\phi| \vec{x})
& \geq &
\log \frac{1}{S} \sum_{j=1}^{S} \frac{
    p_{\theta}(\phi | \vec{x}, \vec{z}^{(j)}) \,  p_{\theta}(\vec{z}^{(j)} | \vec{x})
}{
    q_{\theta}(\vec{z}^{(j)} | \vec{x}, \phi)
},\\
\vec{z}^{(j)} & \sim & q_{\theta}(\vec{z}^{(j)} | \vec{x}, \phi)
    \qquad j=1,\dots,S.
\end{eqnarray}

{\bf Simplified CVAE.} In our experiments we also investigate a variant of the aforementioned model where $ p_{\theta}(\vec{z} | \vec{x}) = q_{\theta}(\vec{z} | \vec{x}, \phi) = p(z) = \mathcal{N}(0, I)$. Compared to the full CVAE framework, this model, which we refer to as \textit{simplified CVAE} (sCVAE) in the experiments, sacrifices the adaptive input-dependent density of the hidden variable $\vec{z}$ for  faster training and test inference as well as optimization stability. In that case the KL-divergence $KL\big(q_{\theta} \parallel  p_{\theta} \big)$ term in $\hat{\mathcal{L}}_{\textrm{ELBO}}$ becomes zero, and we train for a Monte Carlo estimated log-likelihood of the data:

\begin{eqnarray}
\hat{\mathcal{L}}_{\textrm{sCVAE}}(\theta | \vec{X}, \Phi)
& = & \frac{1}{N} \sum_{i=1}^N \log{\Big( \frac{1}{S} \sum_{j=1}^S p_{\theta}(\phi^{(i)} | \vec{x^{(i)}}, \vec{z}^{(j)}) \Big)}, \\
\!\!\vec{z}^{(j)} & \sim& p(\vec{z}) = \mathcal{N}(0, I), j=1,\dots,S.
\label{eq:scvae_objective}
\end{eqnarray}

In some applications it is necessary to make a single best guess about the pose, that is, to summarize the posterior $p(\phi|\vec{x})$ to a single point prediction $\hat{\phi}$. We now discuss an efficient way to do that.

\subsection{Point Prediction}
\label{subsec:pointwise_predictions}

To obtain an optimal single point prediction we utilize Bayesian decision theory~\cite{berger1980bayesian,premachandran2014empirical,bouchacourt2016disco}
and minimize the expected loss,
\begin{equation}
\hat{\phi}_{\Delta} = \argmin_{\phi \in [0,2\pi)} \,
    \mathbb{E}_{\phi' \sim p(\phi|\vec{x})}\left[\Delta(\phi,\phi')\right],
    \label{eq:decision}
\end{equation}
where $\Delta: [0,2\pi) \times [0,2\pi) \to \mathbb{R}_+$ is a loss function.
We will use the $\Delta_{\textrm{AAD}}(\phi,\phi')$ loss which measures the absolute angular deviation (AAD).
To approximate~(\ref{eq:decision}) we use the empirical approximation of~\cite{premachandran2014empirical} and draw $S$ samples
$\{\phi_j\}$ from $p_{\theta}(\phi | \vec{x})$.
We then use the empirical approximation
\begin{equation}
\hat{\phi}_{\Delta} = \argmin_{j=1,\ldots,S} \, \frac{1}{S}
    \sum_{k=1}^S \Delta(\phi_j, \phi_{k}).
\end{equation}

We now evaluate our models both in terms of uncertainty as well as in terms of point prediction quality.

\section{Experiments}
\label{sec:experiments}
%


This section presents the experimental results on several challenging head and object pose regression tasks. Section \ref{subsec:net_arch} introduces the experimental setup including used datasets, network architecture and training setup. In Section \ref{subsec:results} we present and discuss  qualitative and quantitative results on the datasets of interest. 

\subsection{Experimental Setup}
\label{subsec:net_arch}

{\bf Network architecture.} We build our probabilistic framework on top of the Biternion network approach. Therefore, for all the experiments on the head pose regression, we use the same deep batch-normalized VGG-style network~\cite{simonyan2014very} architecture as in~\cite{beyer2015biternion}. The architecture consists of six convolutional layers with 24, 24, 48, 48, 64 and 64 feature channels, respectively, followed by a fully-connected layer of a variable length. The final layer was set to match the number of required parameters of the probabilistic model. 
For the CVAE, the same architecture is used for both the encoder and decoder network. 

For the experiments on the PASCAL3D+ dataset \cite{xiang_wacv14} we use InceptionResNet \cite{szegedy2017inception} as a backbone architecture and jointly predict distributions over three angles (azimuth, elevation and tilt). We use a separate model for each class of objects and consider constructing  a single shot probabilistic object detector and pose estimator a future work.
All models were implemented in Keras~\cite{chollet2015keras} using the TensorFlow~\cite{abadi2016tensorflow} back-end. 
Code and data for all experiments are available at \url{https://github.com/sergeyprokudin/deep_direct_stat}.

{\bf Training.}  We optimize using Adam~\cite{kingma2014adam} and perform random search proposed in~\cite{bergstra2012random} for finding the best values of hyper-parameters such as dropout values, batch size, fully connected layers sizes, learning rate, and other optimizer parameters. 
For the headpose estimation tasks we train all networks for 1000 epochs, with an early stopping in case of no improvement of validation loss after 200 consecutive step. For PASCAL3D+ we train for 200 epochs with an early stopping after 10 epochs of no improvement.

{\bf Head pose datasets.} We evaluate all methods together with the non-probabilistic BiternionVGG baseline on three diverse (in terms of image quality and precision of provided ground truth information) headpose datasets: IDIAP head pose~\cite{idiaphead}, TownCentre~\cite{benfold2011stable} and CAVIAR~\cite{fisher2005caviar} coarse gaze estimation. 
The IDIAP head pose dataset contains 66295 head images stemmed from a video recording of a few people in a meeting room. 
Each image has a complete annotation of a head pose orientation in form of pan, tilt and roll angles. 
We take 42304, 11995 and 11996 images for training, validation, and testing, respectively. 
The TownCentre and CAVIAR datasets present a challenging task of a coarse gaze estimation of pedestrians based on low resolution images from surveillance camera videos. 
In case of the CAVIAR dataset, we focus on the part of the dataset containing occluded head instances (hence referred to as CAVIAR-o in the literature). 
We use (10802, 5444, 5445) and (6916, 874, 904) images for the training-validation-testing split for the CAVIAR and TownCentre datasets, respectively.

{\bf PASCAL3D+ object pose dataset.}   The Pascal 3D+ dataset \cite{xiang_wacv14} consists of images from the Pascal \cite{everingham2010pascal} and ImageNet \cite{deng2009imagenet} datasets that have been
labeled with both detection and continuous pose annotations
for the 12 rigid object categories that appear in Pascal
VOC12 \cite{everingham2010pascal} train and validation set. With nearly 3000 object instances per category, this dataset  provide a rich testbed to study general object pose estimation. In our experiments on
this dataset we follow the same protocol as in \cite{su2015render,tulsiani2015viewpoints} for viewpoint estimation: we use ground truth detections for both training and testing, and use Pascal validation set to evaluate and compare the quality of our predictions.

\begin{table*}[t]
\centering
\caption{Quantitative results on the IDIAP head pose estimation dataset \cite{idiaphead} for the three head rotations pan, roll and tilt. In the situation of fixed camera pose, lightning conditions and image quality, all methods show similar performance (methods are considered to perform on par when the difference in performance is less than \it{standard error of the mean}).}
\resizebox{\textwidth}{!}{
\begin{tabular}{l|c c||c c||c c}                                                        \hline \hline                                                                                                            
estimated pose component             & \multicolumn{2}{c||}{pan}                                        & \multicolumn{2}{c||}{tilt}                                       & \multicolumn{2}{c}{roll}                                       \\ \hline
                                     & \multicolumn{1}{c}{MAAD} & \multicolumn{1}{c||}{log-likelihood} & \multicolumn{1}{c}{MAAD} & \multicolumn{1}{c||}{log-likelihood} & \multicolumn{1}{c}{MAAD} & \multicolumn{1}{c}{log-likelihood} \\ \hline

Beyer et al. (\cite{beyer2015biternion}), fixed $\kappa$  & $\mathbf{5.8^{\circ} \pm 0.1^* }$    & $0.37 \pm 0.01$                    & $\mathbf{2.4^{\circ} \pm 0.1}$           & $1.31 \pm 0.01$                    & $\mathbf{3.1^{\circ} \pm 0.1}$          & $\mathbf{1.13 \pm 0.01}$                    \\  
  
Ours (single von Mises)                     & $6.3^{\circ} \pm 0.1$            & $\mathbf{0.56\pm0.01}$           & $\mathbf{2.3^{\circ} \pm 0.1}$   & $\mathbf{1.56 \pm 0.01}$          & $3.4^{\circ} \pm 0.1$            & $\mathbf{1.13 \pm 0.01} $                 \\ 
Ours (mixture-CVAE)    & $ 6.4^{\circ} \pm 0.1 $          &  $\approx 0.52 \pm 0.02 $     & $2.9^{\circ} \pm 0.1 $           & $\approx 1.35 \pm 0.01  $                  & $3.5^{\circ} \pm 0.1$            & $\approx 1.05 \pm 0.02$              \\ \hline \hline
\multicolumn{2}{l}{\small{*standard error of the mean (SEM).}}
\end{tabular}
}
\label{table:pose_experiments}
\end{table*}

\begin{table*}[t]
\centering
\caption{Quantitative results on the CAVIAR-o~\cite{fisher2005caviar} and TownCentre~\cite{benfold2011stable} coarse gaze estimation datasets. We see clear improvement in terms of quality of probabilistic predictions for both datasets when switching to mixture models that allow to output multiple hypotheses for gaze direction.}
\resizebox{\textwidth}{!}{
\begin{tabular}{l | c c||c c}
\hline \hline
                                     & \multicolumn{2}{c||}{CAVIAR-o}                       & \multicolumn{2}{c}{TownCentre}                             \\ \hline
                                     & MAAD                    & log-likelihood            & MAAD                     & log-likelihood                   \\ \hline
Beyer et al. (\cite{beyer2015biternion}), fixed $\kappa$ & $5.74^{\circ} \pm 0.13$          & $0.262 \pm 0.031$          & $22.8^{\circ} \pm1 .0 $ &  $-0.89 \pm 0.06$  \\

Ours (single von Mises)                    & $5.53^{\circ} \pm 0.13$          & $0.700 \pm 0.043$          & $22.9^{\circ} \pm 1.1$ & $-0.57\pm0.05$       \\ 
Ours (mixture-finite)             & $\mathbf{4.21^{\circ} \pm 0.16}$ & $\mathbf{1.87 \pm 0.04}$ & $23.5^{\circ} \pm 1.1$ & $\mathbf{-0.50 \pm 0.04} $      \\ 
\hline \hline
\end{tabular}
}
\vspace{-0.5cm}
\label{table:gaze_experiments}
\end{table*}

\begin{table}[t!]
\centering
\caption{Results on PASCAL3D+ viewpoint estimation with ground truth bounding boxes. First two evaluation metrics are defined in \cite{tulsiani2015viewpoints}, where $Acc_{\frac{\pi}{6}}$ measures accuracy (the
higher the better) and $MedErr$ measures error (the lower the better). Additionally, we report the log-likelihood estimation $\log \mathcal{L} $ of the predicted  angles (the higher the better). We can see clear improvement on all metrics when switching to probabilistic setting compared to training for a purely discriminative loss (fixed $\kappa$ case).}
\centering
\resizebox{\textwidth}{!}{
\begin{tabular}{lcccccccccccc|c}
\hline
\hline
                        & \textbf{aero} & \textbf{bike} & \textbf{boat} & \textbf{bottle} & \textbf{bus} & \textbf{car} & \textbf{chair} & \textbf{table} & \textbf{mbike} & \textbf{sofa} & \textbf{train} & \textbf{tv} & \textbf{mean}  \\
$Acc_{\frac{\pi}{6}}$ (Tulsiani et al.\cite{tulsiani2015viewpoints})    &  0.81              &      0.77         &     0.59          &    0.93             &      \textbf{0.98}        &     0.89         &       \textbf{0.80}         &        0.62        &         0.88       &       0.82        &    0.80            &      0.80       &     0.81          \\
$Acc_{\frac{\pi}{6}}$ (Su et al.\cite{su2015render})          &     0.80          &    0.82           &    \textbf{0.62}           &      0.95           & 0.93              &     0.83         &     0.75           &      \textbf{0.86}          &        0.86        &      0.85         &       0.82         &     0.89        &        0.83       \\
$Acc_{\frac{\pi}{6}}$ (Ours, fixed $\kappa$)  & 0.83 & 0.75 & 0.54 & 0.95 & 0.92 & 0.90 & 0.77 & 0.71 & \textbf{0.90} & 0.82 & 0.80  & 0.86   & 0.81          \\
$Acc_{\frac{\pi}{6}}$ (Ours, single v.Mises)  & 0.87 & 0.78 & 0.55 & \textbf{0.97} & 0.95 & \textbf{0.91} & 0.78 & 0.76 & \textbf{0.90} & 0.87 & \textbf{0.84} & \textbf{0.91} & \textbf{0.84} \\
$Acc_{\frac{\pi}{6}}$ (Ours, mixture-sCVAE)       &   \textbf{0.89} &    \textbf{0.83}   & 0.46  & 0.96  &  0.93   & 0.90  &  \textbf{0.80}  &  0.76  &  \textbf{0.90}  & \textbf{0.90}  &  0.82    & \textbf{0.91}   &   \textbf{0.84}         \\
\hline \hline
$MedErr$ (Tulsiani et al.\cite{tulsiani2015viewpoints})    &  13.8             &     17.7          &       21.3        &       12.9          &    5.8          &     9.1         &        14.8       &      15.2          &        14.7        &      13.7         &      8.7          &     15.4        &     13.6          \\
$MedErr$ (Su et al.\cite{su2015render})          &   10.0           &      \textbf{12.5}         &      \textbf{20.0}         &     \textbf{6.7}            &   4.5          &     6.7         &        12.3      &                \textbf{8.6}      &           13.1    &      11.0          &    5.8         &       13.3   &  \textbf{10.4}  \\
$MedErr$ (Ours, fixed $\kappa$)   & 11.4 & 18.1 & 28.1 & 6.9  & 4.0 & 6.6 & 14.6 & 12.1 & 12.9 & 16.4 & 7.0 & 12.9 & 12.6 \\
$MedErr$ (Ours, single v.Mises)       &    \textbf{9.7} &  17.7 & 26.9 & \textbf{6.7} & \textbf{2.7} & 4.9 & 12.5 & 8.7 & 13.2 & 10.0 & 4.7 & \textbf{10.6}  &  10.7       \\
$MedErr$ (Ours, mixture-sCVAE)       &      \textbf{9.7} & 15.5 & 45.6 & \textbf{5.4} & 2.9 & \textbf{4.5} & 13.1 & 12.6 & \textbf{11.8} & \textbf{9.1} & \textbf{4.3} & 12.0  &  12.2     \\
\hline \hline

$\log \mathcal{L} $(Ours, fixed $\kappa$)       &  -0.89  &  \textbf{-0.73}  & -1.21  & 0.18                 &       2.09       &  1.43            &     -0.08           & 0.69               &       \textbf{-0.50}         &        -0.75       &      0.06          &       -1.02      &    $-0.07 \pm 0.15$          \\
$\log \mathcal{L} $(Ours, single v.Mises)    &   0.19  &  -1.12 & -0.30 & 2.40  & \textbf{4.87}  & \textbf{2.85} &  \textbf{0.42}  &  \textbf{0.79}  &  -0.72  &  \textbf{-0.54} & \textbf{2.52}  &  0.52 &    $\mathbf{1.17 \pm 0.07}$   \\
$\log \mathcal{L} $(Ours, mixture-sCVAE)      &   \textbf{0.60} &   \textbf{-0.73} & \textbf{-0.26} & \textbf{2.71} & 4.45 & 2.52 & -0.58 & 0.08 & -0.62 & -0.64 & 2.05 & \textbf{1.14} & $\mathbf{1.15 \pm 0.07}$       \\
\hline \hline
\end{tabular}
}
\label{table:pascal3d}

\end{table}

\subsection{Results and Discussion}
\label{subsec:results}

{\bf Quantitative results.}
We evaluate our methods using both discriminative and probabilistic metrics. We use discriminative metrics that are standard for the dataset of interest in order to be able to compare our methods with previous work. For headpose tasks  we use the mean absolute angular deviation (MAAD), a widely used metric for angular regression tasks. For PASCAL3D+ we use the metrics advocated in \cite{tulsiani2015viewpoints}.
Probabilistic predictions are measured in terms of log-likelihood~\cite{good1952rationaldecisions,gneiting2007scoringrules}, a widely accepted scoring rule for assessing the quality of probabilistic predictions.
We summarize the results in Tables ~\ref{table:pose_experiments}, ~\ref{table:gaze_experiments}  and \ref{table:pascal3d}.
It can be seen from results on IDIAP dataset presented in Table ~\ref{table:pose_experiments} that when camera pose, lightning conditions and  image quality are fixed, all methods perform similarly. In contrast, for the coarse gaze estimation task on CAVIAR we can see a clear improvement in terms of quality of probabilistic predictions for both datasets when switching to mixture models that allow to output multiple hypotheses for gaze direction.
Here low resolution, pure light conditions and presence of occlusions create large diversity in the level of head pose expressions. Finally, on a challenging PASCAL3D+ dataset we can see clear improvement on all metrics and classes  when switching to a probabilistic setting compared to training for a purely discriminative loss (fixed $\kappa$ case). Our methods also show competitive or superior performance compared to state-of-the-art methods on disriminative metrics advocated in \cite{tulsiani2015viewpoints}. Method of \cite{su2015render} uses large amounts of synthesized images in addition to the standard training set that was used by our method. Using this data augmentation technique can also lead to an improved performance of our method and we consider this future work.

\begin{figure*}[t!]
\centering
\includegraphics[width=\textwidth]{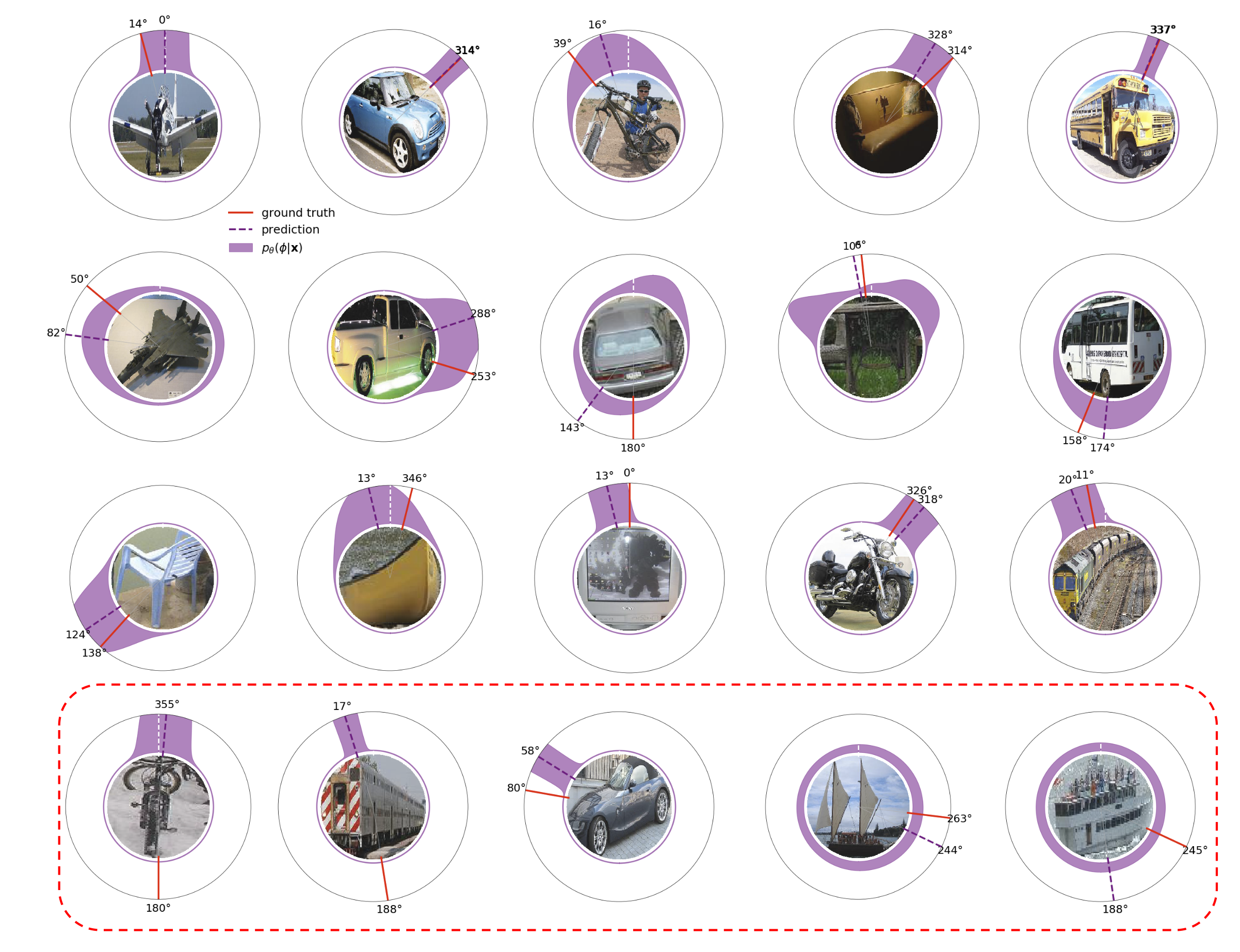}
\caption{Qualitative results of our simpified CVAE model on the PASCAL3D+ dataset. Our model correctly quantifies the uncertainty of pose predictions and is able to model ambiguous cases by predicting complex multimodal densities (second row). Last row shows failure cases (confusing head and tail of the object with high confidence, uniform densities in hard cases).}
\label{fig:qualitative_results}
\end{figure*}

{\bf Qualitative results.} Examples of probabilistic predictions for PASCAL3D+ dataset are shown in Figure \ref{fig:qualitative_results}. The first row highlights the effect we set out to achieve: to correctly quantify the level of uncertainty of the estimated pose. For easier examples we observe sharp peaks and  a highly confident detection, and more spread-out densities otherwise. The examples on the second row highlight the advantage of mixture models, which allow to model complex densities with  multiple peaks corresponding to more than one potential pose angle. Failure scenarios are highlighted in the last row: high confidence predictions in case if the model confuses head and tail of an object and tendency to uniform distributions ($\kappa \xrightarrow{} 0$) for hard classes.

\section{Conclusion}
We demonstrated a new probabilistic model for object pose estimation that is robust to variations in input image quality and accurately quantifies its uncertainty.
More generally our results confirm that our approach is flexible enough to accommodate different output domains such as angular data and enables rich and efficient probabilistic deep learning models. We train all models by maximum likelihood but still find it to be competitive with other works from the literature that explicitly optimize for point estimates even under point estimate loss functions.
In the future, in order to improve our predictive performance and robustness, we would also like to handle uncertainty of model parameters~\cite{kendall2017uncertainties} and to use the Fisher-von Mises distribution to jointly predict a distribution of azimuth-elevation-tilt \cite{mardia2009directional}.

We hope that as intelligent systems increasingly rely on perception abilities, future models in computer vision will be robust and probabilistic.

\clearpage

\section*{Acknowledgments}
    
This work was supported by Microsoft Research through its PhD Scholarship Programme.

\bibliographystyle{splncs}
\bibliography{egbib.bib}

\begin{thebibliography}{10}

\bibitem{marchand2016pose}
Marchand, E., Uchiyama, H., Spindler, F.:
\newblock Pose estimation for augmented reality: a hands-on survey.
\newblock IEEE transactions on visualization and computer graphics
  \textbf{22}(12) (2016)  2633--2651

\bibitem{murphy2009headpose}
Murphy-Chutorian, E., Trivedi, M.M.:
\newblock Head pose estimation in computer vision: A survey.
\newblock IEEE transactions on pattern analysis and machine intelligence
  \textbf{31}(4) (2009)  607--626

\bibitem{poirson2016fast}
Poirson, P., Ammirato, P., Fu, C.Y., Liu, W., Kosecka, J., Berg, A.C.:
\newblock Fast single shot detection and pose estimation.
\newblock In: 3D Vision (3DV), 2016 Fourth International Conference on, IEEE
  (2016)  676--684

\bibitem{massa2016crafting}
Massa, F., Marlet, R., Aubry, M.:
\newblock Crafting a multi-task cnn for viewpoint estimation.
\newblock arXiv preprint arXiv:1609.03894 (2016)

\bibitem{beyer2015biternion}
Beyer, L., Hermans, A., Leibe, B.:
\newblock Biternion nets: Continuous head pose regression from discrete
  training labels.
\newblock In: German Conference on Pattern Recognition, Springer International
  Publishing (2015)  157--168

\bibitem{berger1980bayesian}
Berger, J.O.:
\newblock Statistical Decision Theory and Bayesian Analysis.
\newblock Springer (1980)

\bibitem{xiang2014beyond}
Xiang, Y., Mottaghi, R., Savarese, S.:
\newblock Beyond pascal: A benchmark for 3d object detection in the wild.
\newblock In: Applications of Computer Vision (WACV), 2014 IEEE Winter
  Conference on, IEEE (2014)  75--82

\bibitem{siriteerakul2012advance}
Siriteerakul, T.:
\newblock Advance in head pose estimation from low resolution images: A review.
\newblock International Journal of Computer Science Issues \textbf{9}(2) (2012)

\bibitem{idiaphead}
Odobez, J.M.:
\newblock {IDIAP Head Pose Database.}
\newblock \url{https://www.idiap.ch/dataset/headpose}

\bibitem{gourier2004estimating}
Gourier, N., Hall, D., Crowley, J.L.:
\newblock Estimating face orientation from robust detection of salient facial
  structures.
\newblock In: FG Net Workshop on Visual Observation of Deictic Gestures.
  Volume~6. (2004)

\bibitem{demirkus2014robust}
Demirkus, M., Clark, J.J., Arbel, T.:
\newblock Robust semi-automatic head pose labeling for real-world face video
  sequences.
\newblock Multimedia Tools and Applications \textbf{70}(1) (2014)  495--523

\bibitem{murphy2007head}
Murphy-Chutorian, E., Doshi, A., Trivedi, M.M.:
\newblock Head pose estimation for driver assistance systems: A robust
  algorithm and experimental evaluation.
\newblock In: Intelligent Transportation Systems Conference, 2007. ITSC 2007.
  IEEE, IEEE (2007)  709--714

\bibitem{fisher2005caviar}
Fisher, R., Santos-Victor, J., Crowley, J.:
\newblock Caviar: Context aware vision using image-based active recognition
  (2005)

\bibitem{benfold2011unsupervised}
Benfold, B., Reid, I.:
\newblock Unsupervised learning of a scene-specific coarse gaze estimator.
\newblock In: Computer Vision (ICCV), 2011 IEEE International Conference on,
  IEEE (2011)  2344--2351

\bibitem{fanelli2011real}
Fanelli, G., Gall, J., Van~Gool, L.:
\newblock Real time head pose estimation with random regression forests.
\newblock In: Computer Vision and Pattern Recognition (CVPR), 2011 IEEE
  Conference on, IEEE (2011)  617--624

\bibitem{chamveha2013head}
Chamveha, I., Sugano, Y., Sugimura, D., Siriteerakul, T., Okabe, T., Sato, Y.,
  Sugimoto, A.:
\newblock Head direction estimation from low resolution images with scene
  adaptation.
\newblock Computer Vision and Image Understanding \textbf{117}(10) (2013)
  1502--1511

\bibitem{chen2012we}
Chen, C., Odobez, J.M.:
\newblock We are not contortionists: Coupled adaptive learning for head and
  body orientation estimation in surveillance video.
\newblock In: Computer Vision and Pattern Recognition (CVPR), 2012 IEEE
  Conference on, IEEE (2012)  1544--1551

\bibitem{Flohr2015APF}
Flohr, F., Dumitru-Guzu, M., Kooij, J.F.P., Gavrila, D.:
\newblock A probabilistic framework for joint pedestrian head and body
  orientation estimation.
\newblock IEEE Transactions on Intelligent Transportation Systems \textbf{16}
  (2015)  1872--1882

\bibitem{osadchy2007synergistic}
Osadchy, M., Cun, Y.L., Miller, M.L.:
\newblock Synergistic face detection and pose estimation with energy-based
  models.
\newblock Journal of Machine Learning Research \textbf{8}(May) (2007)
  1197--1215

\bibitem{dantone2012real}
Dantone, M., Gall, J., Fanelli, G., Van~Gool, L.:
\newblock Real-time facial feature detection using conditional regression
  forests.
\newblock In: Computer Vision and Pattern Recognition (CVPR), 2012 IEEE
  Conference on, IEEE (2012)  2578--2585

\bibitem{zhu2012face}
Zhu, X., Ramanan, D.:
\newblock Face detection, pose estimation, and landmark localization in the
  wild.
\newblock In: Computer Vision and Pattern Recognition (CVPR), 2012 IEEE
  Conference on, IEEE (2012)  2879--2886

\bibitem{lu2013headpose}
Lu, J., Tan, Y.P.:
\newblock Ordinary preserving manifold analysis for human age and head pose
  estimation.
\newblock IEEE Transactions on Human-Machine Systems \textbf{43}(2) (2013)
  249--258

\bibitem{huang2011supervised}
Huang, D., Storer, M., De~la Torre, F., Bischof, H.:
\newblock Supervised local subspace learning for continuous head pose
  estimation.
\newblock In: Computer Vision and Pattern Recognition (CVPR), 2011 IEEE
  Conference on, IEEE (2011)  2921--2928

\bibitem{tosato2013characterizing}
Tosato, D., Spera, M., Cristani, M., Murino, V.:
\newblock Characterizing humans on riemannian manifolds.
\newblock IEEE Transactions on Pattern Analysis and Machine Intelligence
  \textbf{35}(8) (2013)  1972--1984

\bibitem{benabdelkader2010robust}
BenAbdelkader, C.:
\newblock Robust head pose estimation using supervised manifold learning.
\newblock Computer Vision--ECCV 2010 (2010)  518--531

\bibitem{geng2014headpose}
Geng, X., Xia, Y.:
\newblock Head pose estimation based on multivariate label distribution.
\newblock In: Proceedings of the IEEE Conference on Computer Vision and Pattern
  Recognition. (2014)  1837--1842

\bibitem{kazemi2014one}
Kazemi, V., Sullivan, J.:
\newblock One millisecond face alignment with an ensemble of regression trees.
\newblock In: Proceedings of the IEEE Conference on Computer Vision and Pattern
  Recognition. (2014)  1867--1874

\bibitem{ba2004probabilistic}
Ba, S.O., Odobez, J.M.:
\newblock A probabilistic framework for joint head tracking and pose
  estimation.
\newblock In: Pattern Recognition, 2004. ICPR 2004. Proceedings of the 17th
  International Conference on. Volume~4., IEEE (2004)  264--267

\bibitem{demirkus2014probabilistic}
Demirkus, M., Precup, D., Clark, J.J., Arbel, T.:
\newblock Probabilistic temporal head pose estimation using a hierarchical
  graphical model.
\newblock In: European Conference on Computer Vision, Springer (2014)  328--344

\bibitem{kendall2017uncertainties}
Kendall, A., Gal, Y.:
\newblock What uncertainties do we need in bayesian deep learning for computer
  vision?
\newblock arXiv preprint arXiv:1703.04977 (2017)

\bibitem{savarese07iccv}
Savarese, S., Fei-Fei, L.:
\newblock 3d generic object categorization, localization and pose estimation.
\newblock In: Computer Vision, 2007. ICCV 2007. IEEE 11th International
  Conference on, IEEE (2007)  1--8

\bibitem{ozuysal2009epflcars}
Ozuysal, M., Lepetit, V., Fua, P.:
\newblock Pose estimation for category specific multiview object localization.
\newblock In: 2009 IEEE Conference on Computer Vision and Pattern Recognition.
  (2009)  778--785

\bibitem{xiang_wacv14}
Xiang, Y., Mottaghi, R., Savarese, S.:
\newblock Beyond pascal: A benchmark for 3d object detection in the wild.
\newblock In: IEEE Winter Conference on Applications of Computer Vision (WACV).
  (2014)

\bibitem{pepik12eccv}
Pepik, B., Gehler, P., Stark, M., Schiele, B.:
\newblock 3d2pm {--} 3d deformable part models.
\newblock In: Proceedings of the European Conference on Computer Vision (ECCV).
  Lecture Notes in Computer Science, Firenze, Springer (October 2012)  356--370

\bibitem{pepik12teaching}
Pepik, B., Stark, M., Gehler, P., Schiele, B.:
\newblock Teaching 3d geometry to deformable part models.
\newblock In: IEEE Conference on Computer Vision and Pattern Recognition
  (CVPR), Providence, RI, USA, IEEE (June 2012)  3362 --3369 oral presentation.

\bibitem{su2015render}
Su, H., Qi, C.R., Li, Y., Guibas, L.J.:
\newblock Render for cnn: Viewpoint estimation in images using cnns trained
  with rendered 3d model views.
\newblock In: Proceedings of the IEEE International Conference on Computer
  Vision. (2015)  2686--2694

\bibitem{tulsiani2015viewpoints}
Tulsiani, S., Malik, J.:
\newblock Viewpoints and keypoints.
\newblock In: Proceedings of the IEEE Conference on Computer Vision and Pattern
  Recognition. (2015)  1510--1519

\bibitem{simonyan2014very}
Simonyan, K., Zisserman, A.:
\newblock Very deep convolutional networks for large-scale image recognition.
\newblock arXiv preprint arXiv:1409.1556 (2014)

\bibitem{szegedy2017inception}
Szegedy, C., Ioffe, S., Vanhoucke, V., Alemi, A.A.:
\newblock Inception-v4, inception-resnet and the impact of residual connections
  on learning.
\newblock In: AAAI. Volume~4. ~12

\bibitem{mardia2009directional}
Mardia, K.V., Jupp, P.E.:
\newblock Directional statistics. Volume 494.
\newblock John Wiley \& Sons (2009)

\bibitem{kingma2013auto}
Kingma, D.P., Welling, M.:
\newblock Auto-encoding variational bayes.
\newblock arXiv preprint arXiv:1312.6114 (2013)

\bibitem{rezende2014stochastic}
Rezende, D.J., Mohamed, S., Wierstra, D.:
\newblock Stochastic backpropagation and approximate inference in deep
  generative models.
\newblock arXiv preprint arXiv:1401.4082 (2014)

\bibitem{sohn2015learning}
Sohn, K., Lee, H., Yan, X.:
\newblock Learning structured output representation using deep conditional
  generative models.
\newblock In: Advances in Neural Information Processing Systems. (2015)
  3483--3491

\bibitem{doersch2016tutorial}
Doersch, C.:
\newblock Tutorial on variational autoencoders.
\newblock arXiv preprint arXiv:1606.05908 (2016)

\bibitem{burda2015iwae}
Burda, Y., Grosse, R., Salakhutdinov, R.:
\newblock Importance weighted autoencoders.
\newblock arXiv preprint arXiv:1509.00519 (2015)

\bibitem{premachandran2014empirical}
Premachandran, V., Tarlow, D., Batra, D.:
\newblock Empirical minimum {Bayes} risk prediction: How to extract an extra
  few \% performance from vision models with just three more parameters.
\newblock In: Proceedings of the IEEE Conference on Computer Vision and Pattern
  Recognition. (2014)  1043--1050

\bibitem{bouchacourt2016disco}
Bouchacourt, D., Mudigonda, P.K., Nowozin, S.:
\newblock {DISCO Nets: DISsimilarity COefficients Networks}.
\newblock In: Advances in Neural Information Processing Systems. (2016)
  352--360

\bibitem{chollet2015keras}
Chollet, F.,  et~al.:
\newblock Keras.
\newblock \url{https://github.com/fchollet/keras} (2015)

\bibitem{abadi2016tensorflow}
Abadi, M., Agarwal, A., Barham, P., Brevdo, E., Chen, Z., Citro, C., Corrado,
  G.S., Davis, A., Dean, J., Devin, M.,  et~al.:
\newblock Tensorflow: Large-scale machine learning on heterogeneous distributed
  systems.
\newblock arXiv:1603.04467 (2016)

\bibitem{kingma2014adam}
Kingma, D., Ba, J.:
\newblock Adam: A method for stochastic optimization.
\newblock arXiv preprint arXiv:1412.6980 (2014)

\bibitem{bergstra2012random}
Bergstra, J., Bengio, Y.:
\newblock Random search for hyper-parameter optimization.
\newblock Journal of Machine Learning Research \textbf{13}(Feb) (2012)
  281--305

\bibitem{benfold2011stable}
Benfold, B., Reid, I.:
\newblock Stable multi-target tracking in real-time surveillance video.
\newblock In: Computer Vision and Pattern Recognition (CVPR), 2011 IEEE
  Conference on, IEEE (2011)  3457--3464

\bibitem{everingham2010pascal}
Everingham, M., Van~Gool, L., Williams, C.K., Winn, J., Zisserman, A.:
\newblock The pascal visual object classes (voc) challenge.
\newblock International journal of computer vision \textbf{88}(2) (2010)
  303--338

\bibitem{deng2009imagenet}
Deng, J., Dong, W., Socher, R., Li, L.J., Li, K., Fei-Fei, L.:
\newblock Imagenet: A large-scale hierarchical image database.
\newblock In: Computer Vision and Pattern Recognition, 2009. CVPR 2009. IEEE
  Conference on, IEEE (2009)  248--255

\bibitem{good1952rationaldecisions}
Good, I.J.:
\newblock Rational decisions.
\newblock Journal of the Royal Statistical Society. Series B (Methodological)
  (1952)  107--114

\bibitem{gneiting2007scoringrules}
Gneiting, T., Raftery, A.E.:
\newblock Strictly proper scoring rules, prediction, and estimation.
\newblock Journal of the American Statistical Association \textbf{102}(477)
  (2007)  359--378

\end{thebibliography}
\end{document}